\pdfoutput=1
\documentclass[11pt]{article}

\usepackage{acl}

\usepackage{times}
\usepackage{latexsym}

\usepackage[T1]{fontenc}
\usepackage[utf8]{inputenc}

\usepackage{microtype}
\usepackage{graphicx}
\usepackage{algorithm}
\usepackage{algorithmic}
\usepackage{amsmath}
\usepackage{amsfonts}
\usepackage{multirow}
\usepackage{booktabs}
\usepackage{makecell}
\title{Enhancing Pre-trained Models with Text Structure Knowledge \\ for Question Generation}

\author{Zichen Wu, Xin Jia, Fanyi Qu, Yunfang Wu\thanks{\ \  Corresponding author.} \\
Key Laboratory of Computational Linguistics, Ministry of Education, China \\
School of Computer Science, Peking University, China\\
\texttt{\{wuzichen,jemmryx,fanyiqu,wuyf\}@pku.edu.cn}}
\begin{document}
\maketitle
\begin{abstract}
% Neural question generation (QG) is to automatically generate questions from given passage with deep neural networks. 
Today the pre-trained language models achieve great success for question generation (QG) task and significantly outperform traditional sequence-to-sequence approaches. However, the pre-trained models treat the input passage as a flat sequence and are thus not aware of the text structure of input passage. For QG task, we model text structure as answer position and syntactic dependency, and propose answer localness modeling and syntactic mask attention to address these limitations. Specially, we present localness modeling with a Gaussian bias to enable the model to focus on answer-surrounded context, and propose a mask attention mechanism to   
% as a supplement to the vanilla self-attention, which 
make the syntactic structure of input passage accessible in question generation process.
% syntactic structure and answer position.  
%Given a piece of answer span and relevant passage, answer-aware question generation model aims to generate more answer-related questions, which is widely applied in real-world scenarios. However, most of the existing works suffer from two problems:
%One method for this is answer-aware question generation, where we take answer and its context as input and generates a answer-relevant question as output. However, this method may suffer from two problems: 
%(1) The generated interrogative words do not match the answer type. (2) The model copies the context words that are far from and irrelevant to the answer. 
%In order to alleviate these issues, some recent works design complicated model structure and incorporate paragraph information and commonsense knowledge to generate more semantic questions, which are well-performed but costly.
    %which have no available pre-training models and lead to unsatisfactory results. 
%local information around answers and generate more answer-related questions. 
% Our method is an update to the self-attention module and can be applied to any pre-trained generative models.  
%approach that incorporate the knowledge around answer and the position information of answer to pre-training models by guiding the calculation process of self-attention module. 
Experiments on SQuAD dataset show that our proposed two modules improve performance over the strong pre-trained model ProphetNet, and combing them together achieves very competitive results with the state-of-the-art pre-trained model.
\end{abstract}

\section{Introduction}
Question generation (QG) aims to generate questions for a given passage, which is a challenging task and shows great value in many practical applications. For instance, QG helps in building reading comprehension tests for course assessments \citep{kurdi2020systematic};
%and can be regarded as a component in intelligent tutoring systems \citep{kurdi2020systematic}. 
%(1) in educational field, forming good questions are vast for evaluating students’ knowledge and stimulating self-learning. QG can generate assessments for course materials or be used as component in intelligent tutoring systems; 
QG can be an important skill for chatbots to start a conversation with human users \citep{mostafazadeh-etal-2016-generating}; QG can help reduce the human labor for collecting large-scale question-answer datasets to improve question answering systems \citep{duan-etal-2017-question}.  

\begin{figure}[ht]
    \centering
    \includegraphics[width=0.5\textwidth]{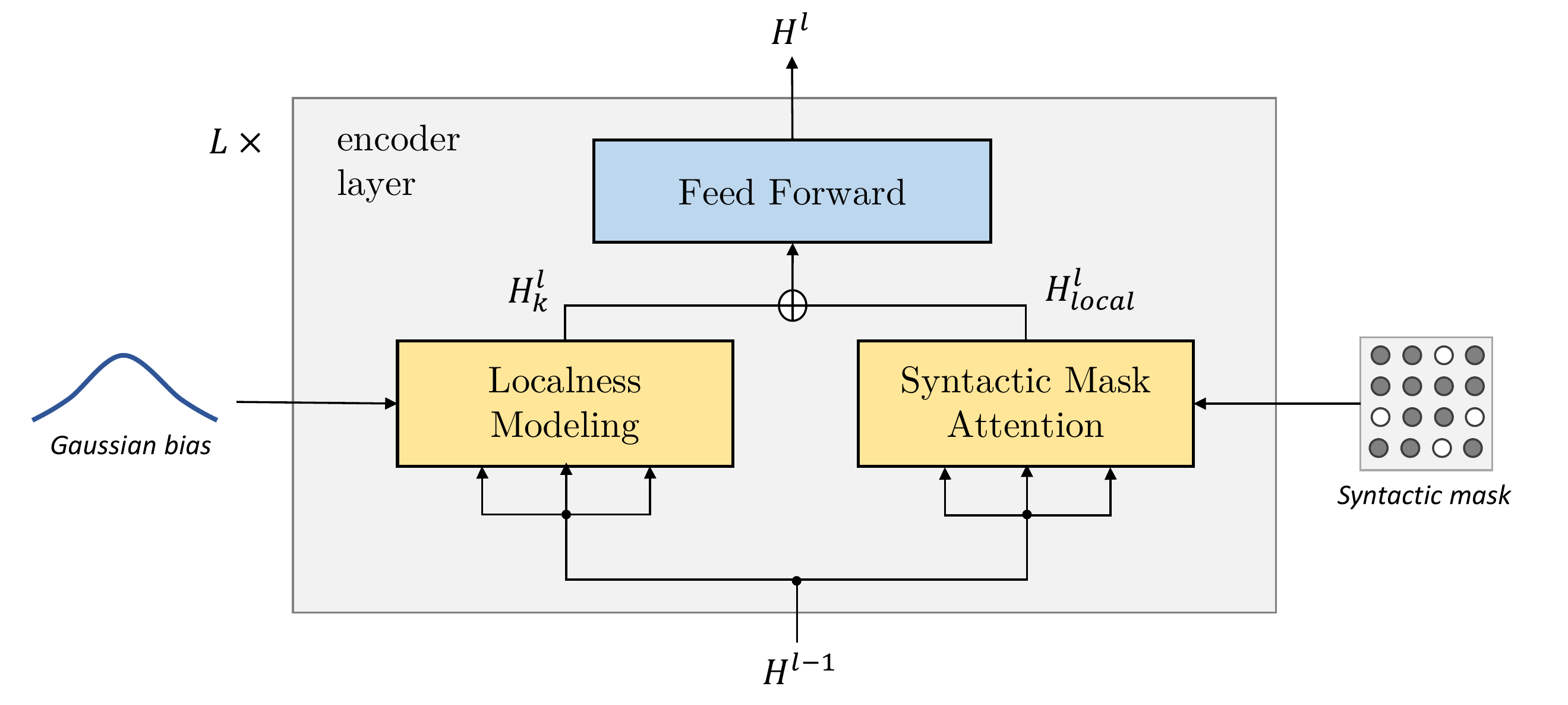}
    \caption{An overall structure of our proposed methods }
    \label{model-fig}
\end{figure}

Existing QG methods mostly employ neural network models by treating it as a sequence-to-sequence generative task, where researchers have tried to incorporate text structure to improve the performance. For answer-aware QG, the answer position is widely explored. 
% answer position and syntactic structure are considered as vital information to generate high-quality answer-aware questions. 
\citet{song-etal-2018-leveraging} propose three strategies to match the answer with passage; \citet{sun-etal-2018-answer} model the relative distance between context words and answer to help generate answer-related questions; \citet{Ma_Zhu_Zhou_Li_2020} enhance model's ability in capturing semantic and positional information via multi-task learning. Some other neural network models dig out deep text structure for generating better questions. For instance, \citet{Chen2020Reinforcement} construct a passage graph to utilize text structure;  \citet{pan-etal-2020-semantic} construct a semantic-level graph to learn the global structure of a document for deep QG. 
% \citet{jia2021enhancing} uses commonsense knowledge to help model generate more human-like questions
However, most of these methods adopt standard neural network, and few of them are based on the strong pre-trained models.
%\citep{du-etal-2017-learning, zhao-etal-2018-paragraph, kim2019improving}, 
% can be roughly categorized into two streams: rule-based and neural network-based methods. Traditional QG methods are mostly rule-based \citep{heilman2009question, heilman-smith-2010-good, labutov-etal-2015-deep, dhole2020syn}, which need human labor to create well-designed templates and patterns, and are thus time-consuming and are of low generalization. Recent QG methods employ neural networks and have made significant progress \citep{du-etal-2017-learning, zhao-etal-2018-paragraph, kim2019improving}, which treat QG as a sequence-to-sequence generative task.
%In these works, QG is tackled as sequence-to-sequence task, where the encoder maps the input sequence into a vector, and the decoder generates the output sequence from hidden vectors. 

Nowadays, pre-trained models largely boost the performance of QG systems \citep{NEURIPS2019_c20bb2d9, bao2020unilmv2, qi-etal-2020-prophetnet, ijcai2020-553}, which utilize huge amounts of unlabeled text corpora in pre-training stage to learn general language representations, and then be fine-tuned in supervised QG tasks. However, in most cases, the input passage and answer are formatted as ``answer [sep] passage'' to feed into the pre-trained model during the fine-tuning stage, as a result that the neural network cannot explicitly capture the syntactic structure of sentence and is not aware of the answer position \citep{zhou2017neural}, which is vital to generate high-quality answer-aware questions. In this end, how to incorporate text structure into pre-trained models for QG remains an open issue.

% and could easily outperforms one without pre-training, where we only need to fin

% However, given an answer and its associated passage, it is possible to focus on the wrong part of context and generate questions that are not fully related to answer. \citet{sun2018answer} study the generation results of models \citet{zhou2017neural} and found that these generated questions suffer from two major issues: (1) over 30\% questions contain the question words that do not match the answer type; (2) 20\% questions copied the context words that are irrelevant to the answer. They attributed these problems to model's failing to pay enough attention to the answer, which should be a guidance in whole generation process, and not being aware of the positions of context words. We argue that these two issues can be regarded as two aspects of lacking answer information: (1) syntactic information, which plays a vital role in determining the question type, sentence components and so on; (2) position information, which helps model to focus more on answer and its local contents. Consequently, we speculate that incorporating syntactic structure and position information will benefit the generation process.
Fortunately, several previous works have tried to utilize external knowledge and position information for NLP tasks. For instance, \citet{yang-etal-2018-modeling} introduce token localness in attention mechanism for machine translation tasks, and \citet{Liu_Zhou_Zhao_Wang_Ju_Deng_Wang_2020} propose a knowledge-enabled language model (K-BERT) to incorporate domain knowledge for question answering systems. 
% many sequence-to-sequence systems have shown promising performance on generating informative texts. \citet{yang-etal-2018-modeling} model token localness as a Gaussian bias in attention mechanism 
%which can be regarded as a type of positional knowledge 
%and achieve better performance on various translation tasks. \citet{Liu_Zhou_Zhao_Wang_Ju_Deng_Wang_2020} propose a knowledge-enabled language model (K-BERT), which effectively incorporates domain knowledge %without separating the vector space of knowledge and source sequence 
%by designing a special visible matrix. 
%For QG task, \citet{sun2018answer} model the relative distance between context words and answer to help model generate answer-related questions. \citet{Ma_Zhu_Zhou_Li_2020} enhance model's ability in capturing semantic and positional information by multi-task learning. % \citet{jia2021enhancing} uses commonsense knowledge to help model generate more human-like questions. 
%However, few of them are based on the strong pre-trained models, and how to incorporate knowledge (syntactic structure and answer position) into pre-trained models for QG remains an open issue.

Motivated by these works, we propose a novel strategy to explicitly incorporate syntactic knowledge and answer information to pre-trained models for QG task. The overall architecture is illustrated in Figure \ref{model-fig}. 
% To enhance model's awareness of position information, 
We present a localness modeling by designing a Gaussian bias to regulate the attention calculation, and propose a new attention mechanism to incorporate syntactic structure. Specially, in localness modeling, we adopt answer position as the center of Gaussian distribution and predict window size automatically, in order that the content around answer will be more focused and the range can be adjusted dynamically according to different token length. For syntactic knowledge, 
% we first extract dependency relations of key sentences as supplement knowledge, and then 
we first compute a syntactic vector through syntactic mask attention, which we design to blind the tokens outside of dependency triples, and then constitute a syntactic-aware representation by adding syntactic vector to the original context vector. 
% What's more, our proposed methods can be easily applied to any Transformer-based network and are thus applicable in almost all pre-trained models. 

We base our method on the strong pre-trained model ProphetNet \cite{qi-etal-2020-prophetnet}, which achieves state-of-the-art results on many generation tasks including QG. 
% We base our model on it in order to (1) verify the effectiveness of our model and (2) achieve some state-of-the-art comparative results.
We conduct experiments on SQuAD dataset \citep{du-etal-2017-learning}. The automatic evaluation shows that our model boosts the performance of pre-trained models and achieves comparable state-of-the-art performance. Further human judgement demonstrates that our model produces high-quality questions by capturing syntactic knowledge and answer-surrounded context. 

Since our method is only a modification to the self-attention module, and there is little increase in parameter number, 
one can easily apply our method to other Transformer-based pre-trained models while almost keeping the original speed without any much computation resources in the training stage. Our codes and data will be publicly available at the Github repository.

\paragraph{Contributions:}To summarize, we (1) propose a modified attention mechanism to incorporate syntactic structure and position information, which can be easily applied to any Transformer-based pre-trained language models for QG task; and (2) with selection of syntactic knowledge, our model is able to reach comparable state-of-the-art performance for question generation.

\section{Related Work}

\subsection{Question Generation}
% Question generation (QG) is a subtask of text generation in natural language processing (NLP), which aims to automatically generate questions from a text passage. 
% Previous works on QG can be roughly divided into two streams: rule-based and neural network-based methods. Rule-based approaches usually rely on manufactured linguistic rules to transform the given text to questions \cite{chali-hasan-2015-towards, lindberg-etal-2013-generating, labutov-etal-2015-deep, dhole2020syn}, and then rank over-generated questions by designed features \cite{heilman2009question, heilman-smith-2010-good}. However, these templates need a lot of human labor and are thus time-consuming,
% %Besides, due to the limitations of human designed rules, 
% and the generated questions may lack diversity.
% and cannot fit all types of input text.

Recently, neural question generation approaches have become popular and achieved great success.  \citet{serban-etal-2016-generating} first introduce an encoder-decoder framework with attention mechanism to generate questions for facts from FreeBase.  \citet{du-etal-2017-learning} employ a seq2seq architecture on QG. \citet{zhao-etal-2018-paragraph} tackle the challenge of processing long text inputs with a gated self-attention encoder. \citet{song-etal-2018-leveraging} and \citet{Kim_Lee_Shin_Jung_2019} strength model's ability by leveraging answer information. \citet{Meng_Ren_Chen_Monz_Ma_de_Rijke_2020} propose a two-stage QG model by creating a draft first and then doing refinement. These works adopt sequence-to-sequence generative approaches and yield many good results on QG. Benefiting from the development of pre-trained model, the performance of QG has been significantly improved. \citet{varanasi-etal-2020-copybert} utilize the information from a self-attention module in BERT \citep{devlin-etal-2019-bert} to generate questions. \citet{qi-etal-2020-prophetnet} propose a pre-training model with n-stream self-attention mechanism and achieve notable results in QG. \citet{ijcai2020-553} design a span-by-span generation flow to predict semantically-complete spans consecutively. So far, \citet{bao2020unilmv2} achieve the best result on SQuAD dataset, which employ autoencoding and partially autoregressive modeling as the pre-training task. 

\begin{figure*}[!ht]
    \centering
    \includegraphics[width=\textwidth]{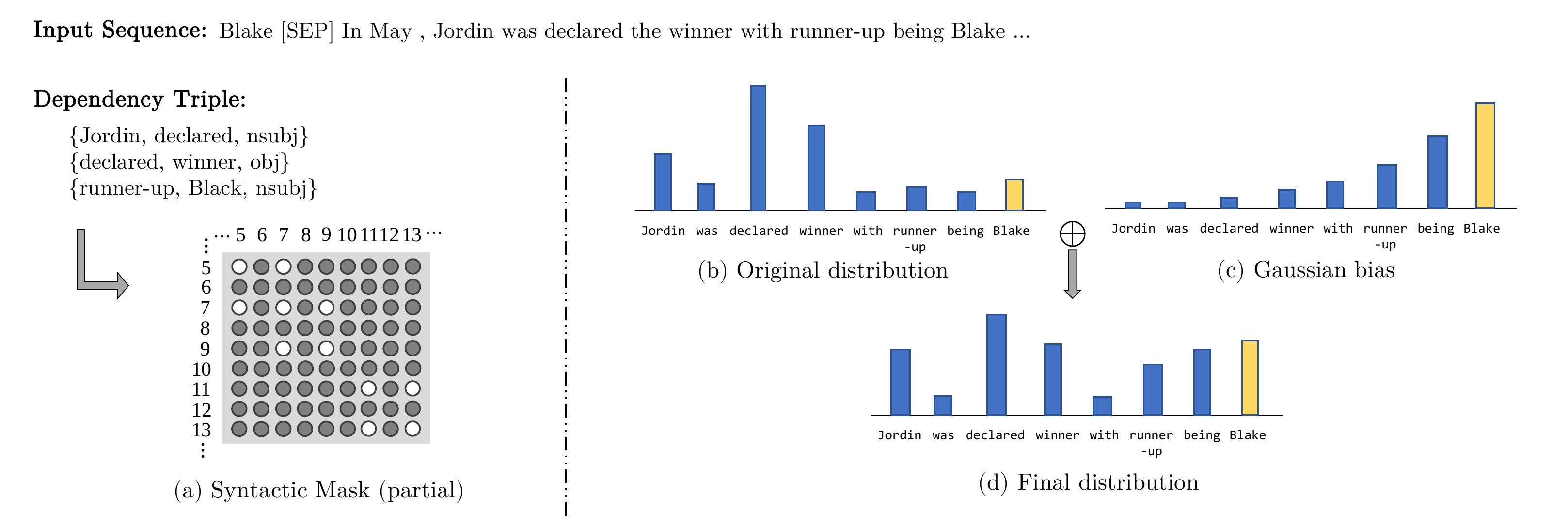}
    \caption{A diagram of Syntactic Mask (left) and Localness Modeling (right). (1) In subfigure (a), we show a partial mask attention with index ranging from 5 to 13 (representing tokens from `Jordin' to `Black' in the input sequence), where the black dot denotes value $-\infty$ and the white denotes 0. (2) Subfigures (b)-(d) show an example of distribution changes in calculating attentions weights of token `declared', where we convert $G$ to the form of distribution for better illustration.}
    \label{localness-fig}
\end{figure*}

\subsection{Knowledge-Enhanced Text Generation}
% In natural language generation tasks, common sense and linguistic knowledge is frequently used to improve the extensiveness of generation models. These knowledge sources can be categorized into internal and external knowledge. Internal knowledge creation takes place within the input texts, including keyword, topic and internal graph structure, while external knowledge is often provided from outside sources, such as knowledge base and grounded text and so on. \citet{yu2020survey} give a good overview of knowledge-enhanced text generation models.

Many knowledge-enhanced text generation systems have achieved great performance on generating informative texts \cite{yu2020survey}. In dialogue systems, topic-aware seq2seq model helps to generate more informative responses \citep{mou-etal-2016-sequence, Xing_Wu_Wu_Liu_Huang_Zhou_Ma_2017, ijcai2018-0643}. In story generation, commonsense knowledge helps to capture the story-line and narrates the plots \citep{Guan_Wang_Huang_2019}.
%step by step, so that the whole story can constitute a reasonable path 
\citet{Liu_Zhou_Zhao_Wang_Ju_Deng_Wang_2020} and \citet{Huang_Fu_Mo_Cai_Xu_Li_Li_Leung_2021} incorporate external knowledge base into the Transformer-like architecture.

For question generation, \citet{zhou2017neural} and \citet{sun-etal-2018-answer} leverage answer embedding and position information to help generate more answer-related questions; \citet{du-cardie-2018-harvesting} incorporate coreference into QG model; \citet{Kim_Lee_Shin_Jung_2019} propose a keyword-net to help the model better capture important information; \citet{jia-etal-2020-ask} integrate paraphrase knowledge into QG systems to generate more human-like questions. 
\citet{Chen2020Reinforcement} and  \citet{pan-etal-2020-semantic} try to capture text structure via graph neural network for QG.  

The attention mechanism has been widely utilized to incorporate knowledge representation, 
%The general idea is to learn a knowledge-aware context vector, %(denoted as $\tilde{c_t}$) 
%by combining both hidden context vector %(denote as $c_t$)
%and knowledge context vector \citep{Meng_Ren_Chen_Monz_Ma_de Rijke_2020, Bi_Wu_Yan_Wang_Xia_Li_2020}, %(denoted as $c_t^K$), such as $\tilde{c_t} = \operatorname{MLP}(c_t \oplus c_t^K)$.
%where the knowledge context vector 
which is obtained by calculating attention over knowledge representations. % such as topic vectors or node vectors in the knowledge graph. 
Multitask training is also an alternative way to incorporate knowledge \citep{Kim2020Sequential, jia-etal-2020-ask}. 

In this paper, based on the pre-trained model for QG, we further incorporate dependency relations and answer position, which can be regarded as structure knowledge. Our task is more challenging
%Our methods are different from these works because we use sentence's syntactic structure as knowledge rather than introducing other knowledge base, 
since 
%our modules are based on the strong pre-trained model, while 
most of the previous works are based on the Transformer only. In order not to increase the computation consumption, we don't use any other sophisticated neural networks beyond the pre-trained model itself.

\section{Preliminaries}
\subsection{Self-Attention}
Here we briefly introduce the encoder of Transformer architecture, on which our methods will base.
% due to its flexibility in parallel computation and its ability to capture long-range dependency.
% high-structure dependency. It calculate attention weights for each pair of token words, thus can capture long-range dependency more directly than RNN counterpart. Formally, 
Given an input sequence $x = \{x_1, x_2, \cdots, x_t\}$, the transformer encoder maps these tokens into hidden representations. More concretely, the $l-1$-th layer hidden states are first linearly transformed into query vectors $q$, key vectors $k$ and value vectors $v$ by multiplying three separate weights,
and then obtain the $l$-th layer hidden states via attention mechanism:
\begin{align}
    H^l &= \operatorname{Attn}(q, k, v) \\
    \operatorname{Attn}(q, k, v) &= \operatorname{softmax} (\frac{scores}{\sqrt{d}})v \\
    scores &= qk^T
\end{align}
where $\sqrt{d}$ is the scaling factor.

% \subsection{Multihead Attention}
%Multihead attention module does this projection $h$ times ($h$ is the number of attention heads) with $h$ different sets of parameters. 
% The multihead attention linearly projects $Q$, $K$, $V$ for $h$ times with $h$ different sets of parameters, in order to jointly attend to information from different representation subspaces at different positions. The process can be defined as:
% \begin{align}
% head_i &= \operatorname{Attention}(Q_i, K_i, V_i)
% \end{align}
% These values are then concatenated and projected to get the final output:
% \begin{multline}
% \operatorname{MultiHead}(Q, K, V) = \\ 
%  \operatorname{Concat}(head_1, \cdots, head_h)W^O
% \end{multline}
% where $W^O$ is the projection weight. 

%Compared to singlehead attention, multihead attention runs through the scaled dot-product attention multiple times in parallel. It allows the model to jointly attend to the information from different representation subspaces at different positions and thus significantly improves the performance of Transformer.

\subsection{Pre-trained Language Model: ProphetNet}
We build our method based on the strong pre-trained model ProphetNet \citep{qi-etal-2020-prophetnet}, which achieves outstanding results on QG tasks. Compared to conventional pre-training models which predict the next one token at each step, ProphetNet tends to predict the next $n$ tokens simultaneously to prevent overfitting on local correlations:
\begin{multline} \label{prophetnet-predict} 
    p(y_t|y_{<t}, x), \cdots, p(y_{t+n-1}|y_{<t},x) = \\ \operatorname{Decoder} (y_{<t}, H_{enc})
\end{multline}
where $H_{enc}$ is the hidden state obtained by the encoder. To achieve this, ProphetNet adopts a n-stream self-attention mechanism to predict next $n$ continuous future tokens, with each stream representing a next $i$-th predicting. 
%The hidden state of each predicting stream is calculated as:
%\begin{align} \label{prophetnet-attention}
%    g_{t-1}^{k+1} = \operatorname{Attention} (g_{t-1}^k, H_{<t}^k \oplus g_{t-1}^k, H_{<t}^k \oplus g_{t-1}^k),
%\end{align}
% where $g_{t-1}^{k}$ represents the $k$-th layer hidden state at step $t-1$.

\section{Integrating Text Structure}

% \subsection{Problem Formulation}
Our work focuses on answer-aware QG task. Formally,
denote the passage, question and answer as $P$, $Q$ and $A$ respectively, the task is defined as:
\begin{align}
    Q^* = \operatorname*{argmax}_{Q} P_{\theta} (Q|P, A)
\end{align}
%where $P, Q$ and $A$ are all composed of tokens. 
Practically, we concatenate passage and answer by format ``$A$ [SEP] $P$'' as the input of ProphetNet, to generate $Q^*$ as output question. A diagram of our method is illustrated in Figure \ref{localness-fig}.

% To tackle this, we present answer localness modeling to guide the model to pay more attention to the content around answer, and propose a new attention mechanism to incorporate syntactic knowledge for QG. 

%called Knowledge Mask Attention for generating syntactic-enhanced questions with answer-passage pairs.
%We denote input token sequence as $x = \{x_1, x_2, \cdots, x_I\}$, where $I$ is the length of the input sequence. Denote the start and end index of the answer span as $s, e$, and the center position of answer is calculated as $(s+e)/2$, which we will use in the following sections. Denote extracted syntactic knowledge as a collection of triples $\varepsilon = (x_i, x_j, r_k)$, where $x_i$ and $x_j$ are the tokens from input sequence, and $r_k$ is the relation between them.

\subsection{Answer Localness Modeling}
In order to enhance model's capability in capturing more local information around the answer and enable the model to generate more answer-related questions, 
%, a preference on answer spans is wanted in our model to generate more answer-related questions. %we want our model to have a preference on answer span, so that generated questions are more related to the answer. From this intuition, 
we model localness as a preference bias to regulate the original attention weight in encoder end, %(Given in Equation (2)-(4)), 
which can be defined as:
\begin{gather}
    \operatorname{Attn}(q, k, v) = \operatorname{softmax}(\frac{scores+G}{\sqrt{d}})v
\end{gather}
where $G \in \mathbb{R}^{I\times I}$ is an alignment matrix, and $I$ refers to the length of input sequence. Each element $G_{i,j}$ represents an alignment preference of token $w_i$ to token $w_j$, which is calculated as:
\begin{gather} \label{G-equation}
    G_{i, j} = -\frac{(j-P_c)^2}{2\sigma_i^2}
\end{gather}
where $P_c$ denotes the central position of answer appearing in the input passage and $\sigma_i$ denotes the standard deviation, which is half of the window size $D_i$. Denote the start and end index of the answer span as $s$ and $e$, then $P_c$ is calculated as:
\begin{gather} \label{fixed-center-equation}
    P_c = \frac{s+e}{2}
\end{gather}

Consequently, after softmax operation, $G$ will become a Gaussian distribution, indicating that the tokens closer to the central position will get higher attention weights.
%the attention scores with an extra addition of a value ranging from $-\infty$ to $0$ will lead to attention weights ranging from $0$ to $1$, and the closer token's position is to the answer, the higher attention weights it will get. 

The window size $D_i$ measures the major area that token $x_i$ needs to align with. We set $D_i$ to be a variable based on the index $i$ rather than a constant value, which we think will bring our model more flexibility to adjust the concentration area according to different token information. %On other words,  
Following \citet{yang-etal-2018-modeling}, we map query vectors into $D_i$ through a feed-forward neural network%, which can be defined as
:
\begin{align}
    z_i &= U_d^T \operatorname{tanh}(W_p q_i)\\
    D_i &= I \cdot \operatorname{sigmoid}(z_i)
\end{align}
where $W_p \in \mathbb{R}^{d\times d}, U_d \in \mathbb{R}^d$ are learnable parameters, and $I$ is a scalar factor regulating $D_i$ to a real value between 0 and the length of input sequence.

Thus, we can obtain hidden states of the input sequence with strengthened localness information:
\begin{align}
    H_{local} = \operatorname{MultiHead}(q,k,v,G)
\end{align}

Figure \ref{localness-fig}(b)-(d) shows an example of answer localness modeling. When calculating the attention weights of token `declared', we first learn a Gaussian distribution centered on the answer `Blake'. The original attention distribution is then regularized to form the final distribution, which not only pays attention to token `winner' but also to token `runner-up' and `Blake'. As a result, the model is guided to attend to the phrase `winner with runner-up being Blake'. 

\subsection{Syntactic Mask Attention}
In this module, we strengthen the syntactic structure of input sentence accessible in question generation process.
%In this paper, we propose a novel attention mechanism to make syntactic knowledge accessible in question generation task. 
The whole procedure for incorporating syntactic knowledge can be divided into three steps: 1) extract appropriate syntactic relations; 2) build syntactic mask and 3) apply syntactic mask to guide the attention calculation. We will introduce them respectively.

\subsubsection{Extract Syntactic Relations} \label{section-extract-relation}
%In order to explicitly introduce syntactic knowledge into the generation model, w
We extract dependency relations and explicitly introduce them into the generation process. Due to the huge time needed for parsing, we only select some key sentences to do syntactic parsing rather than the whole passage. For every answer-passage pairs, %\textless answer, passage, question\textgreater triplet, we first did tokenization on it and split them into sentences. Then 
we select the sentence from the passage where answer spans are located as our key sentence. If the answer spans do not appear completely in any sentences (this is possible because the sentence splitting toolkit is not perfect and sometimes it will mistake punctuation for a terminator), we will select the most similar sentence by computing ROUGE score as an alternative.%, which is a suite of evaluation metrics for automatic text summarization and is used for measuring the similarity between source sentence and target sentence. At last,

Then we apply dependency parser to these selected key sentences, and choose some major relations as the final dependency knowledge, including pred, subj, nsubjpass, csubj, csubjpass, obj, iobj and xcomp. The selection of dependency relations is based on \citet{de-marneffe-etal-2014-universal}. We have also tried some other relations, the analysis of which is given in Section \ref{sec:analysis_dependency}.
%\footnote{The final relations we choose to get main results are ``pred, subj, nsubj, nsubjpass, csubj, csubjpass, obj, iobj, xcomp''}. 
The dependency relation is presented in the form of triples, which can be denoted as $\varepsilon = \{x_i, x_j, r_k\}$, where $x_i$ and $x_j$ are tokens from the input sequence, and $r_k$ denotes the relation between them. 

In our experiment, we utilize the Stanford NLP toolkit \footnote{https://nlp.stanford.edu/software/} to do tokenization, sentence splitting and dependency parsing. A more powerful and sophisticated procedure to extract dependency triples might capture the syntactic structure more accurately, and we leave this to future work.
% as well as enhance the model's awareness to syntactic structure. However, we leave this to future work.
% which contains a set of stable and well-tested natural language processing tools, and is widely used by various groups in academia, industry and government. 

\subsubsection{Syntactic Mask}
Building syntactic mask is the key process in calculating knowledge context vector. Based on the extracted syntactic relation triples, we design a visible mask that prevents some tokens from seeing other ones,  %A syntactic knowledge triple can be denoted as $\varepsilon = <x_i, x_j, r_k>$, where $x_i$ and $x_j$ are the tokens from the input sequence, and $r_k$ is the relation between them. 
which can be defined as below:
\begin{align}
M_{ij}=\left\{
\begin{aligned}
0, & & x_i \ominus x_j \\
-\infty, & & x_i \oslash x_j
\end{aligned}
\right.
\end{align}
where $x_i \ominus x_j$ means $x_i$ and $x_j$ are in a relation triple, while $x_i \oslash x_j$ means they are not.
%For example, in the sentence shown in Figure (),

We display an example in Figure \ref{localness-fig}(a) to show how we construct the syntactic mask from a sentence. We construct the relation triples like $\{\textrm{`Jordin', `declared', `nsubj'}\}$ through extracting dependency relations and build syntactic mask to ensure `Jordin' and `declared' could see each other 
% [???and block their relation with any other tokens with a -$\infty$ mark in the knowledge mask matrix???].
while blind to tokens outside the triples. And tokens not appearing in any triple, like `was', `the' and `being', are not visible to each other.
% e.g., `was', `the' and `being' are not visible to each other. 

%The knowledge mask is defined as below:
%\begin{align}
%M_{ij}=\left\{
%\begin{aligned}
%0 & & x_i \ominus x_j \\
%-\infty & & x_i \oslash x_j
%\end{aligned}
%\right.,
%\end{align}
%where $x_i \ominus x_j$ means $x_i$ and $x_j$ appear in knowledge triples, while $x_i \oslash x_j$ means are not.

\subsubsection{Mask Attention}
The process of mask attention is actually an extension to the original self-attention module in encoder end, whose attention score is calculated as:
% The difference of mask self-attention and vanilla self-attention mechanism appears in the calculation for attention scores:
\begin{align}
scores = qk^T + M
\end{align}
where $M$ is the syntactic mask constructed above.

After the softmax function, the attention score where $M_{ij} = -\infty$ will be 0, indicating that the hidden state of $x_i$ will make no contribution to $x_j$. In this way, we build a syntactic context vector that gathers information of tokens appearing in the extracted syntactic relations while discards unrelated information, which can be a supplementary to the original context vector. Consequently, we obtain the knowledge context vector by:
\begin{align}
    & H_k = \operatorname{MultiHead}(q, k, v, M)
\end{align}

Then the syntactic context vector will be added to the original vector to form the final syntactic-aware representation:
% as a supplementary to form the final vector, the process of
%which can be defined as:
\begin{align}
    & \tilde{H} = H_{local} \oplus H_k
\end{align}
% and the latter uses an additional factor $M$ to fuse syntactic structure information.%Then they are combined to get the knowledge-aware context vector $\tilde{H_c}$.

\subsection{Question Generation}
After obtaining the encoder hidden states $\tilde{H}$, we pass them to the decoder and get the distribution on vocabulary at each step (defined in Equation \ref{prophetnet-predict}). During the training stage, the loss is calculated as:
\begin{align}
    \mathcal{L} = -\sum_{j=0}^{n-1} \alpha_j \left( \sum_{t=1}^{T-j} \log p_\theta (y_{t+j}|y_{<t}, x)\right)
\end{align}
which can be viewed as the weighted sum of $n$ future token predicting loss. And during the inference stage, we disable n-gram predicting and generate a sequence with the highest likelihood:
\begin{align}
    Q = \operatorname*{argmax}_y p_\theta(y_{1:t}|x)
\end{align}
% In real practice, beam search is usually used to find $Q$ in the equation above.

\section{Experimental Setup}
% In this section, we present the experiment details and finetuning results of our approach.

\begin{table*}[t]
\centering
\small
\begin{tabular}{@{}c|c|ccc|ccc@{}}
\toprule[1pt]
\multicolumn{2}{c|}{\multirow{2}{*}{\textbf{Methods (conference-year)}}} & \multicolumn{3}{c|}{\textbf{Du split}}    & \multicolumn{3}{c}{\textbf{Zhao split}}                            \\
\multicolumn{2}{c|}{}                     & BLEU-4         & Meteor         & Rouge-L & BLEU-4               & Meteor               & Rouge-L              \\ \midrule
\multirow{7}{*}{Unpre-trained}           & s2s (ACL-17)                    & 12.28          & 16.62          & 39.75       & -                    & -                    & -                    \\  
% & M2S+cp (NAACL-18)                    & 13.98          & 18.77          & 42.72       & -                    & -                    & -                    \\  
 & CorefNQG (ACL-18)                    & 15.16          & 19.12          & -       & -                    & -                    & -                    \\
 & MP-GSN (EMNLP-18)                      & -              & -              & -       & 16.38                & 20.25                & 44.48     
 \\
%                                        & ASs2s (AAAI-19)                       & 16.20          & 19.92          & 43.96   & -                    & -                    & -                    \\

                                          & SemQG (EMNLP-19)                       & 18.37          & 22.65          & 46.68   & 20.76                & 24.20                & 48.91                \\
                                          & RefNet (EMNLP-19)                      & -              & -              & -       & 18.16                & 23.40                & 47.14                \\
 & ParaphraseQG (ACL-20)                      & 17.21              & 20.96              & -       & -                & -                & -                \\
                                         
                                          & EGSS (AAAI-21)                        & 18.93          & 22.04          & 47.73   & -                    & -                    & -     
                                          \\ \midrule
\multirow{5}{*}{Pre-trained}               
% & UniLM (NeurIPS-19)                       & 22.12          & 25.06          & 51.07   & 23.75                & 25.61                & 52.04                \\

 & $\text{ERINE-GEN}_{\textit{BASE}}$ (IJCAI-20)            & 22.28          & 25.13          & 50.58   & 23.52                & 25.61                & 51.45                \\

& CopyBert (ACL-20)                    & 22.71          & 24.48          & 51.60   & -                    & -                    & -                    \\

& ProphetNet (EMNLP-20)                  & 23.91          & \textbf{26.60} & 52.26   & 25.80                & \textbf{27.54}       & 53.65       \\
                                          
                                          & UniLM-v2 (ICML-20)                    & \textbf{24.43} & 26.34          & 51.97   & 26.29       & 27.16                & 53.22                \\
                                          \cmidrule(l){2-8} 
                                          & Our model                    & 24.37 & 26.26          & \textbf{52.77}   & \textbf{26.30}       & 27.25                & \textbf{53.87}       \\ \bottomrule[1pt]
\end{tabular}
\caption{Experimental results on SQuAD dataset comparing with previous work}
\label{main-results-table}
\end{table*}

\subsection{Dataset}
We conduct experiments on the widely-used reading comprehension dataset SQuAD \citep{rajpurkar-etal-2016-squad}. It consists of questions posed by crowd workers on Wikipedia articles, and the answer to every question is a span extracted from the given passage. There exist different split sets on SQuAD. Following the work of ProphetNet \citep{qi-etal-2020-prophetnet}, we do experiments on two splits: (1) Du split \cite{du-etal-2017-learning}, which splits SQuAD 1.1 dataset into training, development and test sets, consisting of 75,722, 10,570, 11,877 instances respectively; (2) Zhao split \citep{zhao-etal-2018-paragraph}, which uses the reversed dev-test setup as opposed to the original setup.
% used in \citet{du-etal-2017-learning}.
%The training set is used to finetune the model, and the development and test set to evaluate model's performance. 
Following previous work for QG, we adopt BLEU-4 \citep{papineni-etal-2002-bleu}, Meteor \citep{denkowski-lavie-2014-meteor} and Rouge-L \citep{lin-2004-rouge} as evaluation metrics.

\subsection{Training Details}\label{sec-training-details}
We adopt most of the configurations in ProphetNet released by Microsoft group \footnote{\label{prophetnet-github-footnote}https://github.com/microsoft/ProphetNet}. However, due to computation resource limitation, we reimplemented their codes with some different configurations. We use one NVIDIA GeForce RTX 2080 Ti to support fine-tune, and %During the training stage, the first 512 tokens of the passage are fed to the model. The peak learning rate is $1\times 10^{-5}$ and t
during the training stage we truncate overlong input sentences to 512 tokens \footnote{Since ProphetNet only adopts input strings less than 512 tokens.} and the batch size is set to 12. These might bring some decrease in performance, and we take the new results as our baseline for fair comparison.
%Our model is based on Prophetnet \citep{qi2020prophetnet}. %It's a pre-trained model, which predicts the next n tokens simultaneously based on previous context tokens at each time step, and achieves outstanding results on QG tasks. 
Inspired by \citet{yang-etal-2018-modeling}, who show pre-trained models tend to capture localness information in shallow layers while global in higher layers, we apply localness modeling in the first 4 encoder layers and syntactic attention mask in all 12 encoder layers\footnote{In fact, we also conducted experiments on different number of layers, and the results will be reported in our Github repository.}. 
% and report the results of the two data splits as mentioned previously. 

\subsection{Comparing Methods}
%We compare with the following models:
There is a large number of work for QG on SQuAD dataset. Our focus in this paper is to enhance pre-trained models with text structure, so we mainly compare our method with other pre-trained language models. Also, we list some notable sequence-to-sequence works for reference, which did experiments on Du split or Zhao split.   
% We implemented the following model settings to compare:

\textbf{The unpre-trained models for QG.} s2s \citep{du-etal-2017-learning}: an attention-based seq2seq framework for QG. %\textbf{M2S+cp} \citep{song-etal-2018-leveraging}: matching answer and passage to derive answer-aware representations; 
CorefNQG \citep{du-cardie-2018-harvesting}: incorporating coreference knowledge into seq2seq model. MP-GSN \citep{zhao-etal-2018-paragraph}: %proposes a maxout pointer mechanism with gated self-attention encoder and
addressing the challenge of processing long text inputs. %\textbf{ASs2s} \citep{kim2019improving}: an answer-separated seq2seq model;
SemQG \citep{zhang-bansal-2019-addressing}: introducing two semantic-enhanced rewards to regularize generation. RefNet \citep{nema-etal-2019-lets}: a two stage model which creates an initial draft first and then refine it. ParaphraseQG \citep{jia-etal-2020-ask}: incorporating paraphrase knowledge into question generation by back-translation. EGSS \citep{Huang_Fu_Mo_Cai_Xu_Li_Li_Leung_2021}: an entity guided question generation model.
    % with contextual structure information and sequence information capturing;

\textbf{The pre-trained models applied to QG.} ERNIE-GEN \citep{ijcai2020-553}: using multi-granularity target sampling in pre-training and span-by-span generation flow in predicting stage. CopyBert \citep{varanasi-etal-2020-copybert}: utilizing information from self-attention modules of BERT in generation. ProphetNet \citep{qi-etal-2020-prophetnet}: our baseline model, as described before. UniLM-v2 \citep{bao2020unilmv2}: pre-trained of bi-directional language modeling via auto-encoding and seq2seq generation via partially autoregressive modeling.

\begin{table*}[t]
\centering
\small
\begin{tabular}{@{}l|cccccc@{}}
\toprule[1pt]
\multicolumn{1}{l|}{\multirow{2}{*}{\textbf{Method}}} & \multicolumn{3}{c|}{\textbf{Du split}}               & \multicolumn{3}{c}{\textbf{Zhao split}} \\
\multicolumn{1}{c|}{}        & BLEU-4 & Meteor & \multicolumn{1}{c|}{Rouge-L} & BLEU-4 & Meteor & Rouge-L \\ \midrule

ProphetNet*                  & 23.91  & 25.95  & \multicolumn{1}{c|}{52.28}   & 25.71  & 26.99  & 53.63   \\
ProphetNet + Syntactic Mask & \underline{24.22}  & 26.24  & \multicolumn{1}{c|}{\underline{52.60}}   & \underline{26.20}  & 27.01  & 53.74   \\
ProphetNet + Localness      & 24.11  & 26.01  & \multicolumn{1}{c|}{52.52}   & 25.88  & \underline{\textbf{27.28}}  & 53.62   \\
ProphetNet + Syn. Mask + Localness          & \underline{\textbf{24.37}} & \underline{\textbf{26.26}} & \multicolumn{1}{c|}{\underline{\textbf{52.77}}} & \underline{\textbf{26.30}} & \underline{27.25}    & \textbf{53.87} 
\\ \bottomrule[1pt]
\end{tabular}
\caption{Ablation study results by applying different modules on top of the pre-trained ProphetNet model. We report the mean over 3 random seeds. *We re-implemented the ProphetNet released code\textsuperscript{\ref{prophetnet-github-footnote}} and the results are a little lower than the original paper. Underline represents the value is better than baseline with significance ($p < 0.05$).}
\label{ablation-table}
\end{table*}

\section{Results and Analysis}
\subsection{Main Results}

The main experimental results are shown in Table \ref{main-results-table}. For QG on SQuAD dataset, there exists a significant performance gap between the unpre-trained and pre-trained models. The best pre-trained model, UniLM-v2, achieves a 24.43 BLEU-4 on Du split, which receives 5.5 point improvement compared with the best seq2seq model EGSS. The ProphetNet model achieves a competitive BLEU-4 with UniLM-v2, and yields the best result on Meteor and Rouge-L among existing pre-trained models. 
%This is a great confirmation of its capability on generation tasks.%  and we have the following observations:\begin{itemize}\item 

%It's worth noting that our baseline models is a little bit worse than the reported result on the same data, but this shortage can be compensated if we incorporate syntactic and position knowledge into it, which we think proves the efficiency of our methods. 
%\end{itemize}

Considering the excellent performance of these pre-trained models, it's exciting to see that our proposed method with syntactic mask and localness modeling brings further improvement over the basic ProphetNet model, and obtains state-of-the-art results on Zhao split and a close performance with UniLM-v2 on Du split. It's worth noting that our model yields best results in terms of Rouge-L on both datasets among all existing works. 

\subsection{Ablation Study \& Analysis}

\paragraph{Ablation Study} We conduct experiments by applying syntactic mask and localness modeling separately to study their effects on the baseline model. There is a tiny decrease between the result reported by \citet{qi-etal-2020-prophetnet} and our reproduced result on some metrics, which we think is reasonable since we adopt some different configurations forementioned in Section \ref{sec-training-details}. As illustrated in Table \ref{ablation-table}, both components separately enhance the performance of basic ProphetNet and obtain better scores on almost all metrics on two data splits. The performance is further improved when combing two modules together, with a 0.56 BLEU-4 increase on Du split and 0.59 on Zhao split. It indicates that syntactic relations and position information are from two different feature subspace and are able to complement each other in enhancing model's awareness of text structure. And this combining improvement is guaranteed by conducting significance test. % Compared with the basic ProphetNet, our model obtains a 0.56 BLEU-4 increase on Du split and 0.59 on Zhao split
\paragraph{Analysis on Dependency Relations} \label{sec:analysis_dependency}
We conduct experiments to investigate the effect of different dependency relations.  
% First we integrate different kinds of dependency relation into the knowledge vectors to see their effects. 
According to \citet{de-marneffe-etal-2014-universal}, the dependency relations can be divided into two categories: core arguments and non-core arguments, and core arguments can be further divided into nominal relations and clause relations. Inspired by this, we adopt three strategies to select dependency relations: 1) using all dependency relations, which is equal to mask any other tokens except for those in the key sentence; 2) using core arguments, as in Section \ref{section-extract-relation}, which keeps the stem of sentence and gets rid of attributive words; 3) only using nominal relations in core arguments, which removes the clause relation of the stem and thus is the totally plain trunk of the sentence. The choices of core arguments are based on \citet{de-marneffe-etal-2014-universal}.

The experimental results are shown in Table \ref{depen-select-table}. All three strategies bring performance gain over the basic ProphetNet, which validates the effectiveness of our syntactic mask attention. Especially, strategy 2) achieves the best results on both data splits, which provides appropriate dependency knowledge. In contrast, too much dependency relation incorporation may divert the sentence from its correct meaning, which can be found in Strategy 1), and the lack of dependency structure cannot produce a marked effect, which Strategy 3) exemplifies.
\begin{table}[t]
\centering
\small
\begin{tabular}{@{}l|c|c@{}}
\toprule[1pt]
\textbf{Dependency Relations} & \multicolumn{1}{c|}{\textbf{Du split}} & \textbf{Zhao split} \\ \midrule
ProphetNet                    & 23.81                         &   25.71         \\
+ all relations               & 23.93                         &   26.17       \\
+ core arguments              & 24.22                         &   26.20           \\
+ core nominal relations      & 24.01                         &   26.13    \\ \bottomrule[1pt]
\end{tabular}
\caption{Experimental results of different dependency relations in the syntactic mask attention module. }
\label{depen-select-table}
\end{table}
\begin{table}[t]
\centering
\small
\begin{tabular}{@{}l|c|c@{}}
\toprule[1pt]
\textbf{Methods}             & \textbf{Du split} & \textbf{Zhao split}\\ \midrule
ProphetNet          &   23.81   &   25.71  \\
+ predicting center &   23.77   &   25.72  \\
+ answer center     &   24.11   &   25.88  \\ \bottomrule[1pt]
\end{tabular}
\caption{Experimental results of different center strategy in the localness modeling module.}
\label{predict-center-tabel}
\end{table}

\begin{table}[t]
%\centering
\small
\begin{tabular}{@{}clccc@{}}
\toprule[1pt]
\multicolumn{2}{c}{} & \textbf{Base} & \textbf{Ours} & $\rho$ \\ \midrule
\multicolumn{1}{c|}{\multirow{4}{*}{Fluency}}   & \multicolumn{1}{l}{No}             & 4.33\%        & \multicolumn{1}{c|}{4.67\%}  & \multirow{4}{*}{\begin{tabular}[c]{@{}c@{}}0.428\\ (3.8e-06)\end{tabular}} \\
\multicolumn{1}{c|}{}                           & \multicolumn{1}{l}{Med.} & 10.33\%       & \multicolumn{1}{c|}{9.33\%}  &                                                                            \\
\multicolumn{1}{c|}{}                           & \multicolumn{1}{l}{Yes}            & 85.33\%       & \multicolumn{1}{c|}{86\%}    &                                                                            \\ \cmidrule(lr){2-4}
\multicolumn{1}{c|}{}                           & \multicolumn{1}{l}{Avg.}        & 2.81          & \multicolumn{1}{c|}{2.81}    &                                                                            \\ \midrule
\multicolumn{1}{c|}{\multirow{4}{*}{Relevancy}} & \multicolumn{1}{l}{No}             & 1.67\%        & \multicolumn{1}{c|}{1\%}     & \multirow{4}{*}{\begin{tabular}[c]{@{}c@{}}0.507\\ (2.4e-07)\end{tabular}} \\
\multicolumn{1}{c|}{}                           & \multicolumn{1}{l}{Med.}         & 14.33\%       & \multicolumn{1}{c|}{13.67\%} &                                                                            \\
\multicolumn{1}{c|}{}                           & \multicolumn{1}{l}{Yes}            & 84\%          & \multicolumn{1}{c|}{85.33\%} &                                                                            \\ \cmidrule(lr){2-4}
\multicolumn{1}{c|}{}                           & \multicolumn{1}{l}{Avg.}        & 2.82          & \multicolumn{1}{c|}{2.84}    &                                                                            \\ \midrule
\multicolumn{1}{c|}{\multirow{4}{*}{Answerability}} & \multicolumn{1}{l}{No}             & 4\%           & \multicolumn{1}{c|}{3\%}     & \multirow{4}{*}{\begin{tabular}[c]{@{}c@{}}0.456\\ (1.8e-06)\end{tabular}} \\
\multicolumn{1}{c|}{}                           & \multicolumn{1}{l}{Med.}         & 12\%          & \multicolumn{1}{c|}{10.67\%} &                                                                            \\
\multicolumn{1}{c|}{}                           & \multicolumn{1}{l}{Yes}            & 84\%          & \multicolumn{1}{c|}{86.33\%} &                                                                            \\ \cmidrule(lr){2-4}
\multicolumn{1}{c|}{}                           & Avg.                             & 2.80          & \multicolumn{1}{c|}{2.83}    &                                                                            \\ \bottomrule[1pt]
\end{tabular}
\caption{Human evaluation results on the generated questions by ProphetNet (``Base'' ) and our full model (``Ours'' ). Avg. represents the average score of 100 samples. The last column is the Spearman coefficients with p-values in the parentheses.}
\label{human-eval-table}
\end{table}

\paragraph{Analysis on Localness Modeling}
Further, we conduct experiments on different methods to set center position of Gaussian distribution in localness modeling: (1) following \citet{yang-etal-2018-modeling}, who predict the central position by applying a feed-forward transformation to query vector
\begin{gather}
    p_i = U_p^T \operatorname{tanh}(W_p q_i)
\end{gather}
where $W_p$ is a set of parameter to learn; (2) adopt the position of answer as center, as Equation (\ref{fixed-center-equation}). 

The experimental results are shown in Table \ref{predict-center-tabel}. Automatically predicting center position doesn't bring performance gain, while using answer center gets better results on both data splits. We argue that this is because mapping query vector to predict center will lead one token to align with other tokens that are similar with it, which is not applicable in our task. For answer-aware QG, we need each token to pay enough attention to the context around answer, so that the model will obtain better representations with answer as guidance.

\paragraph{Analysis on ProphetNet} One may notice the contradiction between the goal of ProphetNet and our methods: ProphetNet predicts the next several tokens simutaneously to prevent overfitting on local correlations, however, our methods enhance the local syntactic structure in representing stage. Actually, they work on different parts of the pretrained model. ProphetNet uses n-gram prediction strategy in decoder end, while we enhance locality information in encoder end. The enhanced representations of input text, which guides the decoder to generate tokens with fully considering local structure information of the answer, can still benefit from n-gram prediction strategy, since the model would dynamically focus more on neighbour information when previous tokens appearing in the selected structure triples, while keep relatively low information when not. Therefore, more than ProphetNet, our methods should also contributes to other pretrained models.
\subsection{Human Evaluation}
In addition to automatic evaluation, we also evaluate the quality of generated questions by eliciting human judgements. We randomly select 100 $\{\textrm{passage, question, answer}\}$ samples generated by ProphetNet baseline model and our full model, and asked 3 college students to evaluate them. They are required to annotate $yes(3), no(1)$ or $medium(2)$ for each sample from the following aspects: (1) fluency, whether the generated question is grammatical and fluent; (2) relevancy, whether the question is semantically relevant to the input context and (3) answerability, whether the answer is valid to the generated question based on the context. 
    %The option $understandable$ is selected if the question is not totally grammatically correct but we can understand its meaning.
    %Option $yes, no$ or $medium$ represents they are fully, scarcely, or mediocre relevant.% Works will select $yes, no$ or $medium$, where the last one represents the we can only partially answer the question.

The evaluation results are listed in Table \ref{human-eval-table}. %We present the average Spearman coefficients by calculating mean of pairwise Spearman coefficients of the three annotation results. 
The pre-trained model ProphetNet provides us a strong baseline, where most of the generated questions are satisfying. Our model performs better in relevancy and answerability. We speculate this for our proposed method help model capture the information around answer and makes generated question more concentrated to answer. We also reports the average Spearman's coefficients between the annotations, which could guarantee the credibility of our human evaluation results.
% First, we can observe that results of baseline model are already good, which demonstrates the high quality of questions generated by ProphetNet. Furthermore, our model outperforms baseline in relevancy and answerability. We speculate this for that our model has better ability in capturing the information around answer due to our proposed method and thus makes generated question more concentrated to answer. We also reports the Spearman's coefficients between annotators on three indicators, of which the high value guarantee the credibility of our human evaluation results. 

% \begin{figure}[!ht]
%     \centering
%     \includegraphics[width=0.45\textwidth]{Figures/case_study.pdf}
%     \caption{Two examples of generated questions, where words highlighted are answers to those questions.}
%     \label{case-study-fig}
% \end{figure}

\subsection{Case Study}
\begin{table}[t]
\small
\centering
\begin{tabular}{@{}l@{}}
\toprule
\multicolumn{1}{p{0.45\textwidth}}{\textbf{Passage:} ...Meanwhile, a cargo train carrying 13 petrol tanks derailed in Hui County, Gansu, and caught on fire after the \textcolor{red}{\textit{rail was distorted}}.}\\
\multicolumn{1}{l}{\textbf{Answer:} rail was distorted} \\
\multicolumn{1}{l}{\textbf{Generated Questions:}} \\
\multicolumn{1}{l}{\textbf{Golden:} why did the train catch fire?}\\
\multicolumn{1}{l}{\textbf{Base:} why did the cargo train derail in hui county?} \\
\multicolumn{1}{l}{\textbf{Ours:} what caused the cargo train to catch on fire?}
 \\ \midrule
\multicolumn{1}{p{0.45\textwidth}}{\textbf{Passage:} ...A Japanese family with Malaysian citizenship and their 5-year-old child who unfurled a Tibetan flag were hit by a group of Chinese nationals with plastic air-filled batons and heckled by a crowd of Chinese citizens during the confrontation at Independence Square where the relay began, and the Chinese group shouted: "\textcolor{red}{\textit{Taiwan and Tibet belong to China.}}" Later during the day...}\\
\multicolumn{1}{l}{\textbf{Answer:} Taiwan and Tibet belong to China.
} \\
\multicolumn{1}{l}{\textbf{Generated Questions:}} \\
\multicolumn{1}{l}{\textbf{Golden:} what did the chinese group yell?} \\
\multicolumn{1}{l}{\textbf{Base:} what did the chinese say to the japanese family?} \\
\multicolumn{1}{l}{\textbf{Ours:} what did the chinese yell at the japanese family?} \\
\bottomrule
\end{tabular}
\caption{Two examples of generated questions, where answers are highlighted in \textcolor{red}{\textit{Italic font.}}}
\label{table-case-study}
\end{table}

To clearly show the output questions generated by the basic ProphetNet model and our full model, we list two examples in 
Table \ref{table-case-study}.
%we provide questions generated by basic ProphetNet model (base), our full model (ours) and golden question (golden) for the selected SQuAD passage. 
In both examples, the questions generated by our model are closer to golden questions. Specially, in the first example, both the baseline model and our model capture the information that ``rail was distorted'' is the cause to some event. But the event is ``derail'' in base while ``catch on fire'' in ours. The latter is more accurate because ``rail was distorted'' is the direct cause of ``catch on fire'', but the indirect cause of ``derail''. We think the enhanced ability of syntactic information helps capture the sentence structure more accurately and thus find the direct relation. In the second example, both models capture the information that ``Taiwan and Tibet belong to China'' is an utterance said by a Chinese group, but our model focuses more on the answer-surrounded word ``shouted'', leading to generate a more accurate word ``yell''.

% We argue this is because localness modeling makes our model have better ability focusing on the local context. \footnote{We think syntactic knowledge is not the reason because our model will treat ``Taiwan and Tibet belong to China'' as a complete key sentence, no information in other sentences will be strengthened.}
 
\section{Conclusion}
In this work, we enhance the ability of QG systems by incorporating text structure, including syntactic dependency relations and answer position. We strengthen the localness modeling via a learnable Gaussian bias with answer span as center, and present a syntactic mask attention mechanism to fuse structure information. Specifically, we obtain the knowledge-aware context vector by adapting a visible matrix, where each token is only visible to its related token in the knowledge triple. Experimental results on the widely-explored SQuAD dateset demonstrate the effectiveness of our method. 
This work is based on the pre-trained ProphetNet, but our methods can be easily applied to other Transformer-based pre-trained models. In future work, we will validate and expand our method to other NLP tasks, such as summarization, dialog generation, machine reading comprehension, etc.

\section*{Acknowledgement}
This work is supported by the National Natural Science Foundation of  China  (62076008) and the Key Project of Natural Science Foundation of China (61936012).

% These positive results also point to our future work: (1) as our method is not limited to specific tasks, we can validate our method in other tasks, such as reading comprehension, language inference and so on. (2) we can also combine some other techniques in incorporating knowledge or modeling localness to further improve the performance. 

%\section{Acknowledgements}
%We thank the anonymous reviewers for their helpful comments.

\bibliography{anthology,custom}

\begin{thebibliography}{38}
\expandafter\ifx\csname natexlab\endcsname\relax\def\natexlab#1{#1}\fi

\bibitem[{Bao et~al.(2020)Bao, Dong, Wei, Wang, Yang, Liu, Wang, Gao, Piao,
  Zhou et~al.}]{bao2020unilmv2}
Hangbo Bao, Li~Dong, Furu Wei, Wenhui Wang, Nan Yang, Xiaodong Liu, Yu~Wang,
  Jianfeng Gao, Songhao Piao, Ming Zhou, et~al. 2020.
\newblock Unilmv2: Pseudo-masked language models for unified language model
  pre-training.
\newblock In \emph{International Conference on Machine Learning}, pages
  642--652. PMLR.

\bibitem[{Chen et~al.(2020)Chen, Wu, and Zaki}]{Chen2020Reinforcement}
Yu~Chen, Lingfei Wu, and Mohammed~J. Zaki. 2020.
\newblock \href {https://openreview.net/forum?id=HygnDhEtvr} {Reinforcement
  learning based graph-to-sequence model for natural question generation}.
\newblock In \emph{International Conference on Learning Representations}.

\bibitem[{de~Marneffe et~al.(2014)de~Marneffe, Dozat, Silveira, Haverinen,
  Ginter, Nivre, and Manning}]{de-marneffe-etal-2014-universal}
Marie-Catherine de~Marneffe, Timothy Dozat, Natalia Silveira, Katri Haverinen,
  Filip Ginter, Joakim Nivre, and Christopher~D. Manning. 2014.
\newblock \href
  {http://www.lrec-conf.org/proceedings/lrec2014/pdf/1062_Paper.pdf} {Universal
  {S}tanford dependencies: A cross-linguistic typology}.
\newblock In \emph{Proceedings of the Ninth International Conference on
  Language Resources and Evaluation ({LREC}'14)}, pages 4585--4592, Reykjavik,
  Iceland. European Language Resources Association (ELRA).

\bibitem[{Denkowski and Lavie(2014)}]{denkowski-lavie-2014-meteor}
Michael Denkowski and Alon Lavie. 2014.
\newblock \href {https://doi.org/10.3115/v1/W14-3348} {Meteor universal:
  Language specific translation evaluation for any target language}.
\newblock In \emph{Proceedings of the Ninth Workshop on Statistical Machine
  Translation}, pages 376--380, Baltimore, Maryland, USA. Association for
  Computational Linguistics.

\bibitem[{Devlin et~al.(2019)Devlin, Chang, Lee, and
  Toutanova}]{devlin-etal-2019-bert}
Jacob Devlin, Ming-Wei Chang, Kenton Lee, and Kristina Toutanova. 2019.
\newblock \href {https://doi.org/10.18653/v1/N19-1423} {{BERT}: Pre-training of
  deep bidirectional transformers for language understanding}.
\newblock In \emph{Proceedings of the 2019 Conference of the North {A}merican
  Chapter of the Association for Computational Linguistics: Human Language
  Technologies, Volume 1 (Long and Short Papers)}, pages 4171--4186,
  Minneapolis, Minnesota. Association for Computational Linguistics.

\bibitem[{Dong et~al.(2019)Dong, Yang, Wang, Wei, Liu, Wang, Gao, Zhou, and
  Hon}]{NEURIPS2019_c20bb2d9}
Li~Dong, Nan Yang, Wenhui Wang, Furu Wei, Xiaodong Liu, Yu~Wang, Jianfeng Gao,
  Ming Zhou, and Hsiao-Wuen Hon. 2019.
\newblock \href
  {https://proceedings.neurips.cc/paper/2019/file/c20bb2d9a50d5ac1f713f8b34d9aac5a-Paper.pdf}
  {Unified language model pre-training for natural language understanding and
  generation}.
\newblock In \emph{Advances in Neural Information Processing Systems},
  volume~32. Curran Associates, Inc.

\bibitem[{Du and Cardie(2018)}]{du-cardie-2018-harvesting}
Xinya Du and Claire Cardie. 2018.
\newblock \href {https://doi.org/10.18653/v1/P18-1177} {Harvesting
  paragraph-level question-answer pairs from {W}ikipedia}.
\newblock In \emph{Proceedings of the 56th Annual Meeting of the Association
  for Computational Linguistics (Volume 1: Long Papers)}, pages 1907--1917,
  Melbourne, Australia. Association for Computational Linguistics.

\bibitem[{Du et~al.(2017)Du, Shao, and Cardie}]{du-etal-2017-learning}
Xinya Du, Junru Shao, and Claire Cardie. 2017.
\newblock \href {https://doi.org/10.18653/v1/P17-1123} {Learning to ask: Neural
  question generation for reading comprehension}.
\newblock In \emph{Proceedings of the 55th Annual Meeting of the Association
  for Computational Linguistics (Volume 1: Long Papers)}, pages 1342--1352,
  Vancouver, Canada. Association for Computational Linguistics.

\bibitem[{Duan et~al.(2017)Duan, Tang, Chen, and
  Zhou}]{duan-etal-2017-question}
Nan Duan, Duyu Tang, Peng Chen, and Ming Zhou. 2017.
\newblock \href {https://doi.org/10.18653/v1/D17-1090} {Question generation for
  question answering}.
\newblock In \emph{Proceedings of the 2017 Conference on Empirical Methods in
  Natural Language Processing}, pages 866--874, Copenhagen, Denmark.
  Association for Computational Linguistics.

\bibitem[{Guan et~al.(2019)Guan, Wang, and Huang}]{Guan_Wang_Huang_2019}
Jian Guan, Yansen Wang, and Minlie Huang. 2019.
\newblock \href {https://doi.org/10.1609/aaai.v33i01.33016473} {Story ending
  generation with incremental encoding and commonsense knowledge}.
\newblock \emph{Proceedings of the AAAI Conference on Artificial Intelligence},
  33(01):6473--6480.

\bibitem[{Huang et~al.(2021)Huang, Fu, Mo, Cai, Xu, Li, Li, and
  Leung}]{Huang_Fu_Mo_Cai_Xu_Li_Li_Leung_2021}
Qingbao Huang, Mingyi Fu, Linzhang Mo, Yi~Cai, Jingyun Xu, Pijian Li, Qing Li,
  and Ho-fung Leung. 2021.
\newblock \href {https://ojs.aaai.org/index.php/AAAI/article/view/17544}
  {Entity guided question generation with contextual structure and sequence
  information capturing}.
\newblock \emph{Proceedings of the AAAI Conference on Artificial Intelligence},
  35(14):13064--13072.

\bibitem[{Jia et~al.(2020)Jia, Zhou, Sun, and Wu}]{jia-etal-2020-ask}
Xin Jia, Wenjie Zhou, Xu~Sun, and Yunfang Wu. 2020.
\newblock \href {https://doi.org/10.18653/v1/2020.acl-main.545} {How to ask
  good questions? try to leverage paraphrases}.
\newblock In \emph{Proceedings of the 58th Annual Meeting of the Association
  for Computational Linguistics}, pages 6130--6140, Online. Association for
  Computational Linguistics.

\bibitem[{Kim et~al.(2020)Kim, Ahn, and Kim}]{Kim2020Sequential}
Byeongchang Kim, Jaewoo Ahn, and Gunhee Kim. 2020.
\newblock \href {https://openreview.net/forum?id=Hke0K1HKwr} {Sequential latent
  knowledge selection for knowledge-grounded dialogue}.
\newblock In \emph{International Conference on Learning Representations}.

\bibitem[{Kim et~al.(2019)Kim, Lee, Shin, and Jung}]{Kim_Lee_Shin_Jung_2019}
Yanghoon Kim, Hwanhee Lee, Joongbo Shin, and Kyomin Jung. 2019.
\newblock \href {https://doi.org/10.1609/aaai.v33i01.33016602} {Improving
  neural question generation using answer separation}.
\newblock \emph{Proceedings of the AAAI Conference on Artificial Intelligence},
  33(01):6602--6609.

\bibitem[{Kurdi et~al.(2020)Kurdi, Leo, Parsia, Sattler, and
  Al-Emari}]{kurdi2020systematic}
Ghader Kurdi, Jared Leo, Bijan Parsia, Uli Sattler, and Salam Al-Emari. 2020.
\newblock \href {http://www.ijsrp.org/research-paper-0115/ijsrp-p3757.pdf} {A
  systematic review of automatic question generation for educational purposes}.
\newblock \emph{International Journal of Artificial Intelligence in Education},
  30(1):121--204.

\bibitem[{Lin(2004)}]{lin-2004-rouge}
Chin-Yew Lin. 2004.
\newblock \href {https://aclanthology.org/W04-1013} {{ROUGE}: A package for
  automatic evaluation of summaries}.
\newblock In \emph{Text Summarization Branches Out}, pages 74--81, Barcelona,
  Spain. Association for Computational Linguistics.

\bibitem[{Liu et~al.(2020)Liu, Zhou, Zhao, Wang, Ju, Deng, and
  Wang}]{Liu_Zhou_Zhao_Wang_Ju_Deng_Wang_2020}
Weijie Liu, Peng Zhou, Zhe Zhao, Zhiruo Wang, Qi~Ju, Haotang Deng, and Ping
  Wang. 2020.
\newblock \href {https://doi.org/10.1609/aaai.v34i03.5681} {K-bert: Enabling
  language representation with knowledge graph}.
\newblock \emph{Proceedings of the AAAI Conference on Artificial Intelligence},
  34(03):2901--2908.

\bibitem[{Ma et~al.(2020)Ma, Zhu, Zhou, and Li}]{Ma_Zhu_Zhou_Li_2020}
Xiyao Ma, Qile Zhu, Yanlin Zhou, and Xiaolin Li. 2020.
\newblock \href {https://doi.org/10.1609/aaai.v34i05.6366} {Improving question
  generation with sentence-level semantic matching and answer position
  inferring}.
\newblock \emph{Proceedings of the AAAI Conference on Artificial Intelligence},
  34(05):8464--8471.

\bibitem[{Meng et~al.(2020)Meng, Ren, Chen, Monz, Ma, and
  de~Rijke}]{Meng_Ren_Chen_Monz_Ma_de_Rijke_2020}
Chuan Meng, Pengjie Ren, Zhumin Chen, Christof Monz, Jun Ma, and Maarten
  de~Rijke. 2020.
\newblock \href {https://doi.org/10.1609/aaai.v34i05.6370} {Refnet: A
  reference-aware network for background based conversation}.
\newblock \emph{Proceedings of the AAAI Conference on Artificial Intelligence},
  34(05):8496--8503.

\bibitem[{Mostafazadeh et~al.(2016)Mostafazadeh, Misra, Devlin, Mitchell, He,
  and Vanderwende}]{mostafazadeh-etal-2016-generating}
Nasrin Mostafazadeh, Ishan Misra, Jacob Devlin, Margaret Mitchell, Xiaodong He,
  and Lucy Vanderwende. 2016.
\newblock \href {https://doi.org/10.18653/v1/P16-1170} {Generating natural
  questions about an image}.
\newblock In \emph{Proceedings of the 54th Annual Meeting of the Association
  for Computational Linguistics (Volume 1: Long Papers)}, pages 1802--1813,
  Berlin, Germany. Association for Computational Linguistics.

\bibitem[{Mou et~al.(2016)Mou, Song, Yan, Li, Zhang, and
  Jin}]{mou-etal-2016-sequence}
Lili Mou, Yiping Song, Rui Yan, Ge~Li, Lu~Zhang, and Zhi Jin. 2016.
\newblock \href {https://aclanthology.org/C16-1316} {Sequence to backward and
  forward sequences: A content-introducing approach to generative short-text
  conversation}.
\newblock In \emph{Proceedings of {COLING} 2016, the 26th International
  Conference on Computational Linguistics: Technical Papers}, pages 3349--3358,
  Osaka, Japan. The COLING 2016 Organizing Committee.

\bibitem[{Nema et~al.(2019)Nema, Mohankumar, Khapra, Srinivasan, and
  Ravindran}]{nema-etal-2019-lets}
Preksha Nema, Akash~Kumar Mohankumar, Mitesh~M. Khapra, Balaji~Vasan
  Srinivasan, and Balaraman Ravindran. 2019.
\newblock \href {https://doi.org/10.18653/v1/D19-1326} {Let{'}s ask again:
  Refine network for automatic question generation}.
\newblock In \emph{Proceedings of the 2019 Conference on Empirical Methods in
  Natural Language Processing and the 9th International Joint Conference on
  Natural Language Processing (EMNLP-IJCNLP)}, pages 3314--3323, Hong Kong,
  China. Association for Computational Linguistics.

\bibitem[{Pan et~al.(2020)Pan, Xie, Feng, Chua, and
  Kan}]{pan-etal-2020-semantic}
Liangming Pan, Yuxi Xie, Yansong Feng, Tat-Seng Chua, and Min-Yen Kan. 2020.
\newblock \href {https://doi.org/10.18653/v1/2020.acl-main.135} {Semantic
  graphs for generating deep questions}.
\newblock In \emph{Proceedings of the 58th Annual Meeting of the Association
  for Computational Linguistics}, pages 1463--1475, Online. Association for
  Computational Linguistics.

\bibitem[{Papineni et~al.(2002)Papineni, Roukos, Ward, and
  Zhu}]{papineni-etal-2002-bleu}
Kishore Papineni, Salim Roukos, Todd Ward, and Wei-Jing Zhu. 2002.
\newblock \href {https://doi.org/10.3115/1073083.1073135} {{B}leu: a method for
  automatic evaluation of machine translation}.
\newblock In \emph{Proceedings of the 40th Annual Meeting of the Association
  for Computational Linguistics}, pages 311--318, Philadelphia, Pennsylvania,
  USA. Association for Computational Linguistics.

\bibitem[{Qi et~al.(2020)Qi, Yan, Gong, Liu, Duan, Chen, Zhang, and
  Zhou}]{qi-etal-2020-prophetnet}
Weizhen Qi, Yu~Yan, Yeyun Gong, Dayiheng Liu, Nan Duan, Jiusheng Chen, Ruofei
  Zhang, and Ming Zhou. 2020.
\newblock \href {https://doi.org/10.18653/v1/2020.findings-emnlp.217}
  {{P}rophet{N}et: Predicting future n-gram for
  sequence-to-{S}equence{P}re-training}.
\newblock In \emph{Findings of the Association for Computational Linguistics:
  EMNLP 2020}, pages 2401--2410, Online. Association for Computational
  Linguistics.

\bibitem[{Rajpurkar et~al.(2016)Rajpurkar, Zhang, Lopyrev, and
  Liang}]{rajpurkar-etal-2016-squad}
Pranav Rajpurkar, Jian Zhang, Konstantin Lopyrev, and Percy Liang. 2016.
\newblock \href {https://doi.org/10.18653/v1/D16-1264} {{SQ}u{AD}: 100,000+
  questions for machine comprehension of text}.
\newblock In \emph{Proceedings of the 2016 Conference on Empirical Methods in
  Natural Language Processing}, pages 2383--2392, Austin, Texas. Association
  for Computational Linguistics.

\bibitem[{Serban et~al.(2016)Serban, Garc{\'\i}a-Dur{\'a}n, Ahn, Chandar,
  Courville, and Bengio}]{serban-etal-2016-generating}
Iulian~Vlad Serban, Alberto Garc{\'\i}a-Dur{\'a}n, Sungjin Ahn, Sarath Chandar,
  Aaron Courville, and Yoshua Bengio. 2016.
\newblock \href {https://doi.org/10.18653/v1/P16-1056} {Generating factoid
  questions with recurrent neural networks: The 30{M} factoid question-answer
  corpus}.
\newblock In \emph{Proceedings of the 54th Annual Meeting of the Association
  for Computational Linguistics (Volume 1: Long Papers)}, pages 588--598,
  Berlin, Germany. Association for Computational Linguistics.

\bibitem[{Song et~al.(2018)Song, Wang, Hamza, Zhang, and
  Gildea}]{song-etal-2018-leveraging}
Linfeng Song, Zhiguo Wang, Wael Hamza, Yue Zhang, and Daniel Gildea. 2018.
\newblock \href {https://doi.org/10.18653/v1/N18-2090} {Leveraging context
  information for natural question generation}.
\newblock In \emph{Proceedings of the 2018 Conference of the North {A}merican
  Chapter of the Association for Computational Linguistics: Human Language
  Technologies, Volume 2 (Short Papers)}, pages 569--574, New Orleans,
  Louisiana. Association for Computational Linguistics.

\bibitem[{Sun et~al.(2018)Sun, Liu, Lyu, He, Ma, and
  Wang}]{sun-etal-2018-answer}
Xingwu Sun, Jing Liu, Yajuan Lyu, Wei He, Yanjun Ma, and Shi Wang. 2018.
\newblock \href {https://doi.org/10.18653/v1/D18-1427} {Answer-focused and
  position-aware neural question generation}.
\newblock In \emph{Proceedings of the 2018 Conference on Empirical Methods in
  Natural Language Processing}, pages 3930--3939, Brussels, Belgium.
  Association for Computational Linguistics.

\bibitem[{Varanasi et~al.(2020)Varanasi, Amin, and
  Neumann}]{varanasi-etal-2020-copybert}
Stalin Varanasi, Saadullah Amin, and Guenter Neumann. 2020.
\newblock \href {https://doi.org/10.18653/v1/2020.nlp4convai-1.3}
  {{C}opy{BERT}: A unified approach to question generation with
  self-attention}.
\newblock In \emph{Proceedings of the 2nd Workshop on Natural Language
  Processing for Conversational AI}, pages 25--31, Online. Association for
  Computational Linguistics.

\bibitem[{Xiao et~al.(2020)Xiao, Zhang, Li, Sun, Tian, Wu, and
  Wang}]{ijcai2020-553}
Dongling Xiao, Han Zhang, Yukun Li, Yu~Sun, Hao Tian, Hua Wu, and Haifeng Wang.
  2020.
\newblock \href {https://doi.org/10.24963/ijcai.2020/553} {Ernie-gen: An
  enhanced multi-flow pre-training and fine-tuning framework for natural
  language generation}.
\newblock In \emph{Proceedings of the Twenty-Ninth International Joint
  Conference on Artificial Intelligence, {IJCAI-20}}, pages 3997--4003.
  International Joint Conferences on Artificial Intelligence Organization.
\newblock Main track.

\bibitem[{Xing et~al.(2017)Xing, Wu, Wu, Liu, Huang, Zhou, and
  Ma}]{Xing_Wu_Wu_Liu_Huang_Zhou_Ma_2017}
Chen Xing, Wei Wu, Yu~Wu, Jie Liu, Yalou Huang, Ming Zhou, and Wei-Ying Ma.
  2017.
\newblock \href {https://ojs.aaai.org/index.php/AAAI/article/view/10981} {Topic
  aware neural response generation}.
\newblock \emph{Proceedings of the AAAI Conference on Artificial Intelligence},
  31(1).

\bibitem[{Yang et~al.(2018)Yang, Tu, Wong, Meng, Chao, and
  Zhang}]{yang-etal-2018-modeling}
Baosong Yang, Zhaopeng Tu, Derek~F. Wong, Fandong Meng, Lidia~S. Chao, and Tong
  Zhang. 2018.
\newblock \href {https://doi.org/10.18653/v1/D18-1475} {Modeling localness for
  self-attention networks}.
\newblock In \emph{Proceedings of the 2018 Conference on Empirical Methods in
  Natural Language Processing}, pages 4449--4458, Brussels, Belgium.
  Association for Computational Linguistics.

\bibitem[{Yu et~al.(2020)Yu, Zhu, Li, Hu, Wang, Ji, and Jiang}]{yu2020survey}
Wenhao Yu, Chenguang Zhu, Zaitang Li, Zhiting Hu, Qingyun Wang, Heng Ji, and
  Meng Jiang. 2020.
\newblock A survey of knowledge-enhanced text generation.
\newblock \emph{arXiv preprint arXiv:2010.04389}.

\bibitem[{Zhang and Bansal(2019)}]{zhang-bansal-2019-addressing}
Shiyue Zhang and Mohit Bansal. 2019.
\newblock \href {https://doi.org/10.18653/v1/D19-1253} {Addressing semantic
  drift in question generation for semi-supervised question answering}.
\newblock In \emph{Proceedings of the 2019 Conference on Empirical Methods in
  Natural Language Processing and the 9th International Joint Conference on
  Natural Language Processing (EMNLP-IJCNLP)}, pages 2495--2509, Hong Kong,
  China. Association for Computational Linguistics.

\bibitem[{Zhao et~al.(2018)Zhao, Ni, Ding, and Ke}]{zhao-etal-2018-paragraph}
Yao Zhao, Xiaochuan Ni, Yuanyuan Ding, and Qifa Ke. 2018.
\newblock \href {https://doi.org/10.18653/v1/D18-1424} {Paragraph-level neural
  question generation with maxout pointer and gated self-attention networks}.
\newblock In \emph{Proceedings of the 2018 Conference on Empirical Methods in
  Natural Language Processing}, pages 3901--3910, Brussels, Belgium.
  Association for Computational Linguistics.

\bibitem[{Zhou et~al.(2018)Zhou, Young, Huang, Zhao, Xu, and
  Zhu}]{ijcai2018-0643}
Hao Zhou, Tom Young, Minlie Huang, Haizhou Zhao, Jingfang Xu, and Xiaoyan Zhu.
  2018.
\newblock \href {https://doi.org/10.24963/ijcai.2018/643} {Commonsense
  knowledge aware conversation generation with graph attention}.
\newblock In \emph{Proceedings of the Twenty-Seventh International Joint
  Conference on Artificial Intelligence, {IJCAI-18}}, pages 4623--4629.
  International Joint Conferences on Artificial Intelligence Organization.

\bibitem[{Zhou et~al.(2017)Zhou, Yang, Wei, Tan, Bao, and
  Zhou}]{zhou2017neural}
Qingyu Zhou, Nan Yang, Furu Wei, Chuanqi Tan, Hangbo Bao, and Ming Zhou. 2017.
\newblock \href {http://tcci.ccf.org.cn/conference/2017/papers/1084.pdf}
  {Neural question generation from text: A preliminary study}.
\newblock In \emph{National CCF Conference on Natural Language Processing and
  Chinese Computing}, pages 662--671. Springer.

\end{thebibliography}
\bibliographystyle{acl_natbib}

\end{document}